%% file: main.tex
\documentclass[10pt,twocolumn,letterpaper]{article}

\usepackage[pagenumbers]{iccv} 
\usepackage{algorithm}
\usepackage{algorithmic}
\usepackage{tabularx}
\usepackage{float}

\input{preamble}

\definecolor{cvprblue}{rgb}{0.21,0.49,0.74}
\usepackage[pagebackref,breaklinks,colorlinks,allcolors=cvprblue]{hyperref}

\author{%
Hamadi Chihaoui \quad  Paolo Favaro \\
Computer Vision Group, University of Bern, Switzerland\\
\texttt{\{hamadi.chihaoui,paolo.favaro\}@unibe.ch}\\
}

\title{Diffusion Image Prior}

\begin{document}
\maketitle

\begin{abstract}
Zero-shot image restoration (IR) methods based on pretrained diffusion models have recently achieved significant success. These methods typically require at least a parametric form of the degradation model. However, in real-world scenarios, the degradation may be too complex to define explicitly. To handle this general case, we introduce the \methodName (\methodnameacr).
We take inspiration from the Deep Image Prior (DIP) \cite{ulyanov2018deep}, 
since it can be used to remove artifacts without the need for an explicit degradation model. However, in contrast to DIP, we find that pretrained diffusion models offer a much stronger prior, despite being trained without knowledge from corrupted data. 
We show that, the optimization process in \methodnameacr first reconstructs a clean version of the image before eventually overfitting to the degraded input, but it 
does so for a broader range of degradations than DIP. In light of this result, we propose a blind image restoration (IR) method based on early stopping, which does not require prior knowledge of the degradation model. 
We validate  \methodnameacr on various degradation-blind IR tasks, including JPEG artifact removal, waterdrop removal, denoising and super-resolution with state-of-the-art results.
\end{abstract}

\begin{figure}[t]
\centering\
   \setkeys{Gin}{width=0.49\linewidth}
    \captionsetup[subfigure]{skip=0.0ex,
                             belowskip=0.0ex,
                             labelformat=simple}
                             \setlength{\tabcolsep}{1.0pt}
    \renewcommand\thesubfigure{}
    \small

\small
  \begin{tabular}{cc}
 
 {\includegraphics{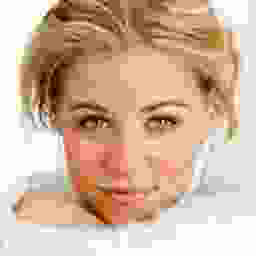}}& {\includegraphics{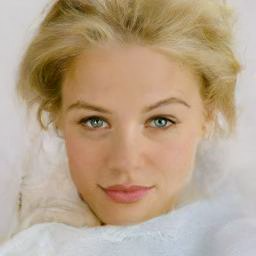}} \\
  \textrm{Input} & \textrm{DreamClean~\cite{xiaodreamclean}} \\
 {\includegraphics{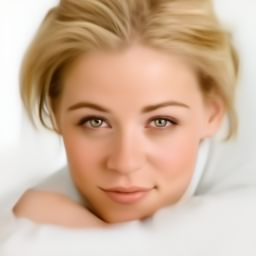}}& {\includegraphics{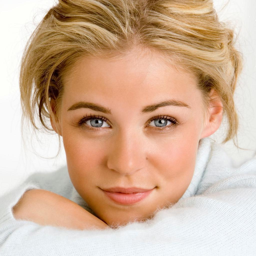}}\\
  \textrm{Ours (\methodnameacr)} &  \textrm{Ground-truth}\\
   \end{tabular}
\captionof{figure}{\label{fig:ood_teaser} In practical applications, image degradation is often unknown, intricate, and challenging to model, such as when removing JPEG artifacts. In comparison to the recent state-of-the-art DreamClean~\cite{xiaodreamclean}, \methodName (\methodnameacr) consistently produces results that more effectively preserve the original identity of the image. \vspace{.5em}}
\end{figure}
\section{Introduction}

One of the most well-known training-free and blind methods for image restoration is Deep Image Prior (DIP) \cite{ulyanov2018deep}. A notable feature of DIP is its applicability in cases where the degradation model is too complex to be accurately modeled, such as with JPEG compression. DIP leverages the implicit prior (bias) embedded within a convolutional neural network (CNN). It demonstrates that through an iterative reconstruction process of a corrupt image (without any knowledge of the degradation model) with an untrained CNN, one can recover a clean image. Inspired by this approach, we explore a similar investigation, where we exploit the reconstruction process of a degraded image, but use instead a frozen pre-trained (diffusion) model.
Although our reconstruction optimizes only the input (noise) to the model, while in DIP the CNN parameters were optimized, we find that the two processes share the same property of reconstructing a clean image before converging to the corresponding degraded input.
Surprisingly, we find that while in DIP this property holds primarily for high frequency degradations such as noise, with pre-trained diffusion models this phenomenon is observed across a wider range of degradations, including low-frequency ones such as blurring. This happens despite the fact that the diffusion model was never exposed to such degraded data during training to either generate it or to be adversarial to it. We call this property the \emph{\methodName} (\methodnameacr). 

\methodnameacr has immediate applicability in image restoration (IR), where the aim is to reconstruct a clean image from its degraded version. While DIP can handle only image corruption processes that inject high-frequency artifacts in an image without using the degradation operator, \methodnameacr can handle a broader range of artifacts. Recently, DreamClean~\cite{xiaodreamclean} has also demonstrated the remarkable capability of restoring images without knowledge of the degradation model for a wide range of cases. However, as shown in Figure~\ref{fig:ood_teaser}, \methodnameacr yields reconstructions that are more faithful to the original image. The capability of the above three methods drastically simplifies the solution of IR tasks, as they do not require any collection of specialized datasets for supervised training \cite{liang2021swinir, wang2021real}, which can be costly and time-consuming, nor the use of additional knowledge about the artifacts \cite{kawar2022denoising, lugmayr2022repaint, chung2022improving, chung2022diffusion,chung2023parallel, fei2023generative, chihaoui2024blind}, which is often not available in practice. 

We demonstrate \methodnameacr on several IR problems, non-uniform deformation, waterdrop removal, different noise distributions (mixtures of Gaussian and speckle noise), JPEG compression and superresolution. Our findings indicate that  in all these degradation cases the reconstruction of the initial noise through a pre-trained and frozen diffusion model initially generates clean reconstructions before overfitting to the degraded images. This suggests that \methodnameacr can be used in a similar fashion to DIP  via \emph{early stopping}. In this paper, we present some  self-supervised early stopping criteria that result in state of the art performance on all the above IR tasks despite their simplicity. 

In summary, our contributions are
\begin{itemize}

    \item We conduct a study showing the inductive biases of a frozen pretrained diffusion model when used for image reconstruction in the case of a wide range of degradations that either remove high-frequency details or introduce high-frequency artifacts; we show that these models exhibit a similar implicit prior behavior as observed with DIP, but on a broader range of cases;
 
    \item Based on our findings, we propose a new training-free and fully blind image restoration method, \methodnameacr, which does not assume any prior knowledge of the degradation model, making it broadly applicable to a wide range of complex image restoration tasks;

    \item We demonstrate state-of-the-art performance on CelebA and ImageNet benchmarks for several blind image restoration tasks, including denoising, waterdrop removal, super-resolution, and JPEG artifact removal.

\end{itemize}

\begin{figure*}[t]
\centering\
   \setkeys{Gin}{width=0.2\linewidth}
    \captionsetup[subfigure]{skip=0.0ex,
                             belowskip=0.0ex,
                             labelformat=simple}
                             \setlength{\tabcolsep}{0.0pt}
    \renewcommand\thesubfigure{}
    \small

\small
  \begin{tabular}{cccccc}
  \textit{Superresolution} & \textit{JPEG De-artifacting} & \textit{Waterdrop removal}& \textit{Denoising}& \textit{Non-uniform deformation} \\
 {\includegraphics{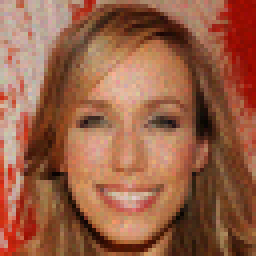}}& {\includegraphics{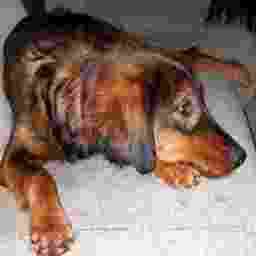}}& 
  {\includegraphics{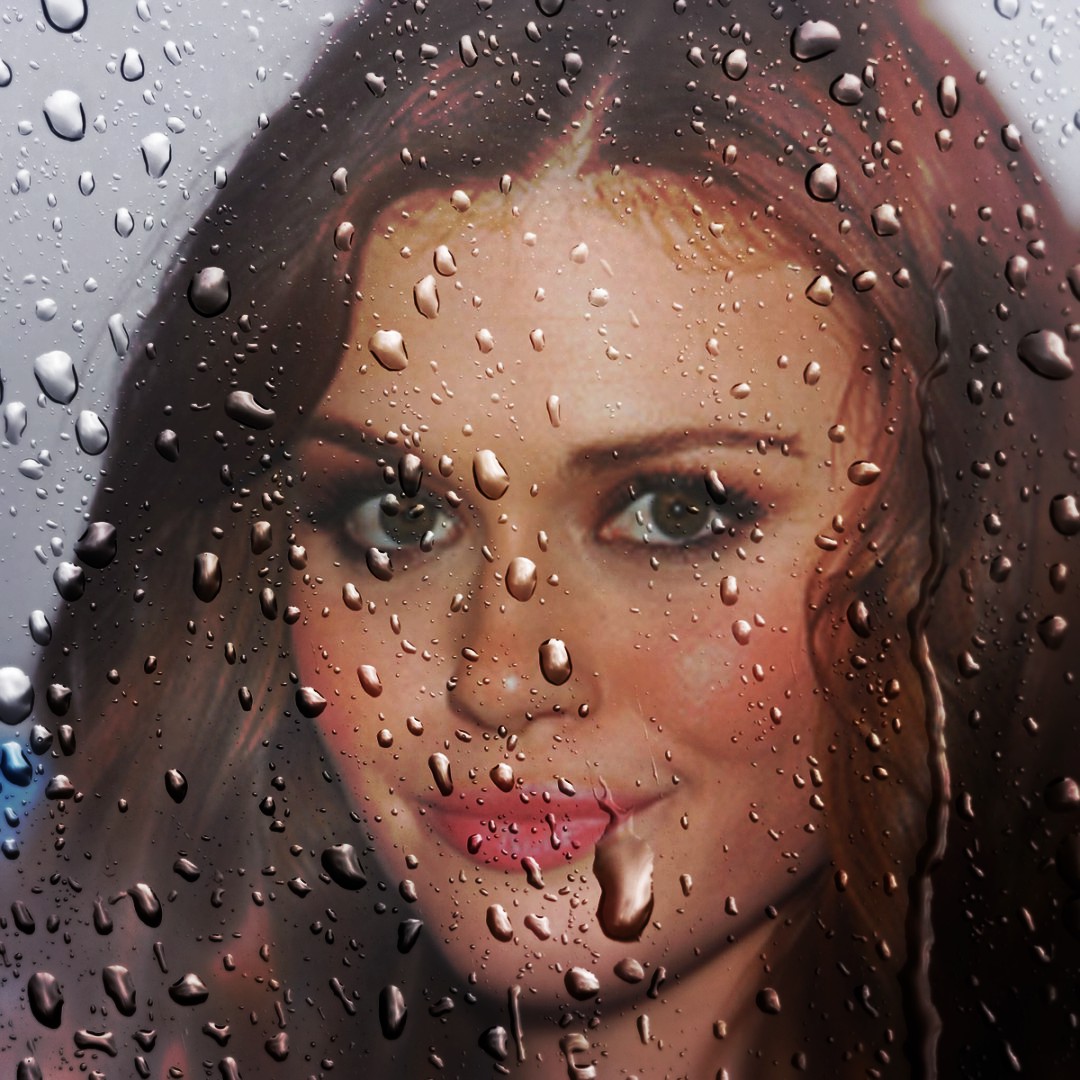}} & 
 {\includegraphics{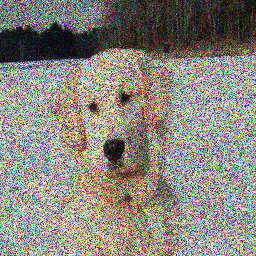}} & 
 {\includegraphics{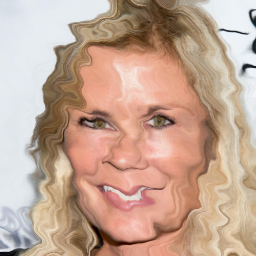}}\\
 {\includegraphics{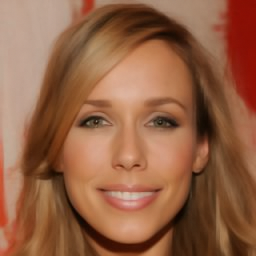}}& {\includegraphics{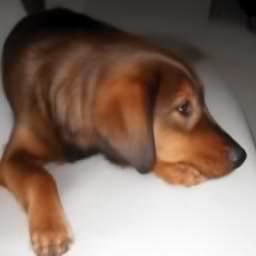}}& 
  {\includegraphics{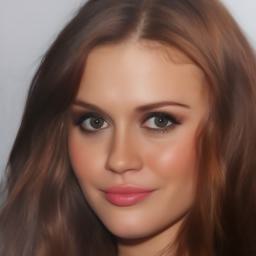}}& 
 {\includegraphics{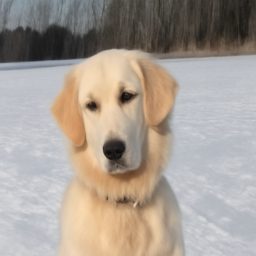}}& 
 {\includegraphics{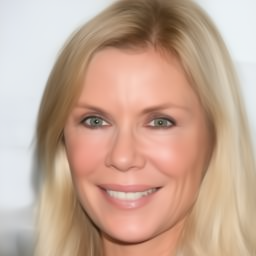}} \\
 {\includegraphics{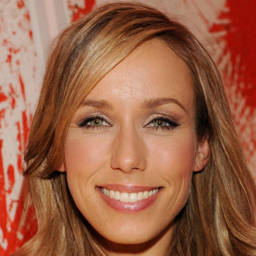}}& {\includegraphics{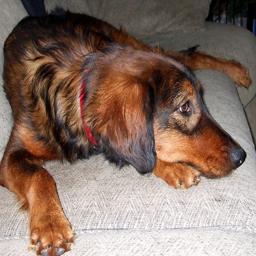}}& 
  {\includegraphics{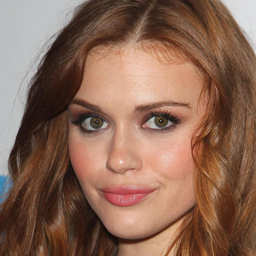}} & 
 {\includegraphics{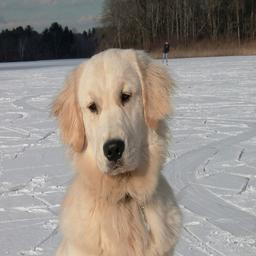}}& 
 {\includegraphics{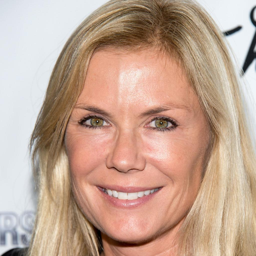}} \\
   \end{tabular}

\captionof{figure}{\label{fig:ood_teaser2} We demonstrate \methodnameacr on several \textbf{degradation-blind} image restoration tasks. Top: Degraded input. Middle: Our prediction. Bottom: Ground truth.\vspace{.5em}}
\end{figure*}

\section{Related Work}

Image restoration has a long and rich history in the literature. In this review, we focus on image restoration methods that are similar to our approach in terms of their training paradigm (test-time training methods) and blindness (fully blind).

\subsection{Training-based Image Restoration}
Training-based image restoration methods \cite{zhang2017beyond, liang2021swinir, zamir2020cycleisp} involve training a model for image restoration on some training dataset. 
SwinIR \cite{liang2021swinir} is a supervised, transformer-based architecture that performs well across various image restoration tasks, including denoising and super-resolution. In the unsupervised domain, CycleISP \cite{zamir2020cycleisp} addresses image super-resolution and restoration by using unpaired images through a cycle-consistent adversarial network. These methods generally require large datasets for training and may suffer from generalization issues when tested on new data that differs from the training set.

\subsection{Test-time Image Restoration}
Recently, test-time training methods have gained popularity. These methods eliminate the need for a training dataset and are applied directly to the degraded image at test time. We can further classify these methods based on their degradation model blindness.

\noindent{\textbf{Non-Blind and Partially-Blind Image Restoration.}}
Non-blind methods assume that the degradation model is fully known; for example, in image deblurring, they assume the degradation takes the form of a blur kernel and that the entries of this kernel are known. Some recent non-blind methods include \cite{kawar2022denoising, chung2022diffusion, wang2022zero}. DDRM \cite{kawar2022denoising} introduces a variational inference objective and proposes an inverse problem solver based on posterior sampling to learn the posterior distribution of the inverse problem. DDNM \cite{wang2022zero} presents a zero-shot framework for image restoration tasks using range-null space decomposition, refining only the null-space contents during the reverse diffusion process to both ensure data fidelity and realism. DPS \cite{chung2022diffusion} offers a more general framework that addresses both non-linear and noisy cases. However, these methods assume complete knowledge of the degradation model, which is rarely available in real-world scenarios.
On the other hand, partially blind methods relax this assumption by only requiring knowledge of the parametric form of the degradation model. These methods include \cite{chihaoui2024blind, murata2023gibbsddrm, fei2023generative, chung2023parallel}. GibbsDDRM \cite{murata2023gibbsddrm} extends DDRM to scenarios where only partial information about the degradation is available. BlindDPS \cite{chung2023parallel} builds on the work of \cite{chung2022diffusion}, addressing blind deblurring by jointly optimizing the blur operator and the sharp image during the reverse diffusion process. BIRD \cite{chihaoui2024blind} proposes a fast diffusion inversion method that simultaneously inverts the diffusion model and infers the degradation model.

\noindent{\textbf{Fully-Blind image Restoration Methods.}
There have been a few attempts to solve image restoration in a zero-shot manner when the explicit parametric form of the degradation is unknown. One such method is Deep Image Prior (DIP), which can handle complex degradation scenarios despite its reliance on the inductive bias of untrained neural networks. Recently, DreamClean \cite{xiaodreamclean}, a fully blind image restoration method based on a pre-trained diffusion model, was introduced. DreamClean works by inverting the input image and iteratively moving the intermediate latents through an additional variance-preserving step during the reverse diffusion sampling process. In this work, we explore a different approach to using a pre-trained diffusion model for the fully blind case. Inspired by DIP, we aim to leverage the power of pre-trained diffusion models without requiring an explicit degradation model, demonstrating how this approach can achieve robust image restoration.

\section{The \methodName}


\subsection{Image Restoration}\label{assumptions}

In this work, we aim to solve the task of image restoration in a zero-shot and fully blind manner. Given an input degraded image $y$, our goal is to recover the degradation-free image $x$. We leverage a pre-trained diffusion model, specifically the DDIM~\cite{song2020denoising}, a class of diffusion models that enables deterministic mapping from noise space to data space. We denote the \textbf{deterministic} mapping that transforms a Gaussian noise vector into a data sample from the target distribution as $g$.   

We focus on degradations that are rarely addressed by existing zero-shot and blind methods, as these degradations are either too complex to model in a zero-shot setting or simply unknown. Examples include noise with unknown distributions (\eg, correlated noise or non-Gaussian/Poisson distributions), JPEG compression artifacts, and non-uniform blur. In such cases, the IR task can be formulated as an energy minimization problem of the form
\begin{align}\label{eq:p1}
    x^* = \arg~\min_{x} E(x; y) + R(x),
\end{align}
where $E(x; y)$ is a data fidelity term that measures how well $x$ matches the observed image $y$, and $R(x)$ is a regularization term that enforces prior knowledge about the distribution of $x$. A classic example for restoring images with Gaussian noise is to use $E(x;y) = \|x-y\|^2$ and the Total Variation (TV) of the image as $R(x)$, which encourages solutions to contain uniform regions \cite{elad2023imagedenoisingdeeplearning}. 



\subsection{Revisiting the Deep Image Prior \label{sect:2.2}}

Deep Image Prior (DIP) addresses the optimization problem in \cref{eq:p1} by introducing an implicit prior, which is captured by a class of neural networks, specifically deep Convolutional Neural Networks (CNNs) and \emph{early stopping}, as follows
\begin{align}\label{eq:p2}
    \theta^* = \arg \min_{\theta } \|f_{\theta}(z) - y\|^2, \quad\text{with}\quad x^* = f_{\theta^*}(z),
\end{align}
where we implicitly let $E(f_{\theta}(z); y)=\|f_{\theta}(z) - y\|^2$.
The minimizer $\theta^*$ is obtained using an iterative optimizer such as gradient descent, and starts from a random initialization of the
parameters of a convolutional neural network $f_{\theta}$ fed a fixed input noise vector $z$ (which is not optimized). The early stopped solution $\theta^*$ is then used to compute the restored image  $x^* = f_{\theta^*} (z)$.

\begin{figure*}
\centering
\begin{subfigure}{.5\textwidth}
 \centering\
   \setkeys{Gin}{width=0.16\linewidth}
    \captionsetup[subfigure]{skip=0.0ex,
                             belowskip=0.0ex,
                             labelformat=simple}
                             \setlength{\tabcolsep}{0.0pt}
    \renewcommand\thesubfigure{a}
    \small

\small
  \begin{tabular}{cccccc}
  {\includegraphics{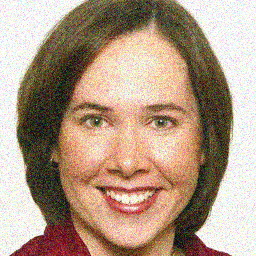}}& 
 {\includegraphics{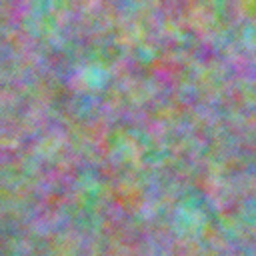}}& 
 {\includegraphics{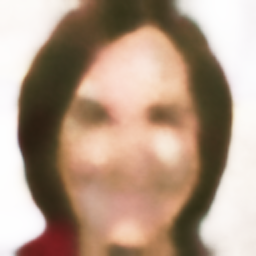}}&{\includegraphics{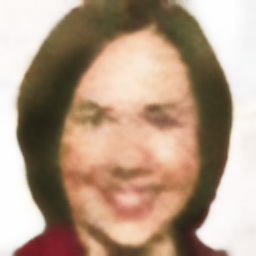}}& 
{\includegraphics{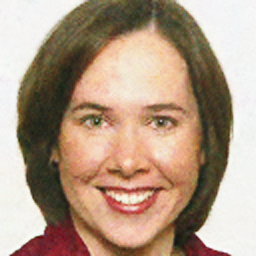}}& 
 {\includegraphics{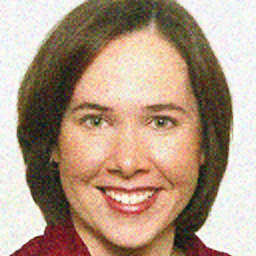}}\\
  {\includegraphics{images/selected_noise/00319.png}}& 
  {\includegraphics{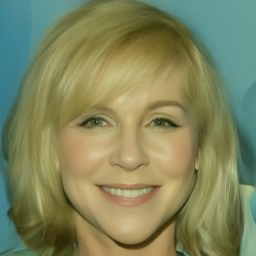}}& {\includegraphics{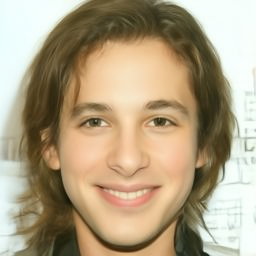}}& 
 {\includegraphics{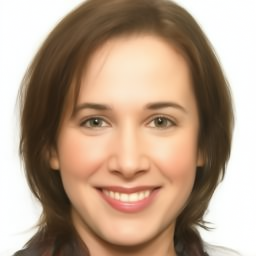}}& 
 {\includegraphics{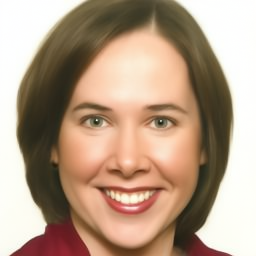}}& 
  {\includegraphics{images/selected_noise/00319.png}}\\
\scriptsize\textit{Input} &  \scriptsize\textit{Iter = 0} &  \scriptsize\textit{Iter = 25} & \scriptsize\textit{Iter = 50}  & \scriptsize\textit{Iter = 500} & \scriptsize\textit{Iter = 1500}\\
   \end{tabular}
\caption{Gaussian noise}
\end{subfigure}%
\begin{subfigure}{.5\textwidth}
 \centering\
   \setkeys{Gin}{width=0.16\linewidth}
    \captionsetup[subfigure]{skip=0.0ex,
                             belowskip=0.0ex,
                             labelformat=simple}
                             \setlength{\tabcolsep}{0.0pt}
    \renewcommand\thesubfigure{b}
    \small

\small
  \begin{tabular}{cccccc}
  {\includegraphics{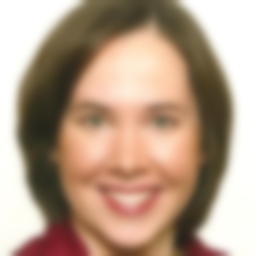}}& 
 {\includegraphics{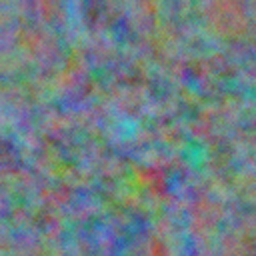}}& 
 {\includegraphics{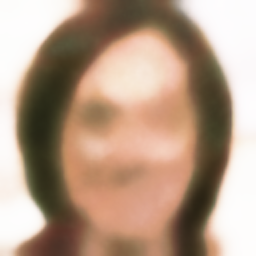}}&{\includegraphics{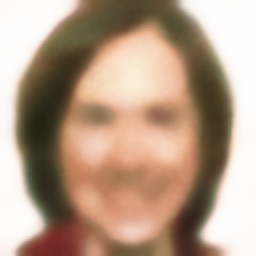}}& 
{\includegraphics{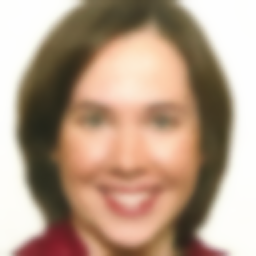}}& 
 {\includegraphics{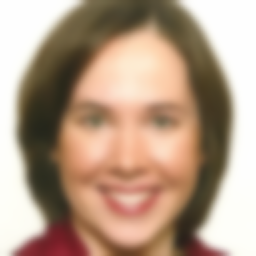}}\\
  {\includegraphics{images/selected_blur/00319.png}}& 
  {\includegraphics{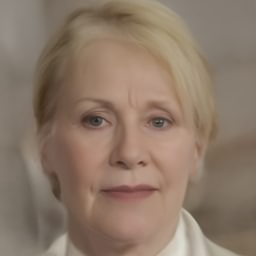}}& {\includegraphics{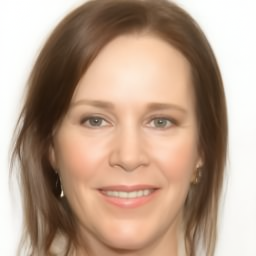}}& 
 {\includegraphics{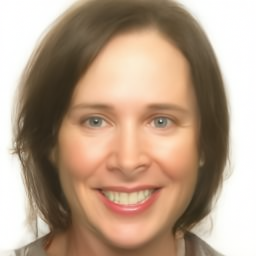}}& 
 {\includegraphics{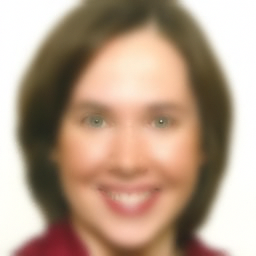}}& 
  {\includegraphics{images/selected_blur/00319.png}}\\
  \scriptsize\textit{Input} &  \scriptsize\textit{Iter = 0} &  \scriptsize\textit{Iter = 25} & \scriptsize\textit{Iter = 50}  & \scriptsize\textit{Iter = 500} & \scriptsize\textit{Iter = 1500}\\
   \end{tabular}
\caption{ Gaussian blur}
  \label{fig:sub2}
\end{subfigure}
\caption{Intermediate outputs of the iterative reconstruction of a degraded image in the case of DIP (top) and our proposed optimization scheme using a frozen pretrained diffusion model (bottom). In the case of a pre-trained diffusion model, the optimization consistently produces clean images at an intermediate stage  \textbf{irrespective of the type of degradation} ((a) noise or (b) blur).}
\label{fig:intermediate-steps}
\end{figure*}

\subsection{Biases of a Pretrained Diffusion Model}

Here, we aim to address two key questions
\begin{enumerate}[Q1:]
    \item Does a pre-trained diffusion model exhibit similar biases to those observed in DIP?
    \item If such biases exist, how do they differ from those in DIP? 
\end{enumerate}
To investigate these questions, we explore the task of reconstructing a degraded image using a pre-trained diffusion model, where $g$ represents its induced mapping from noise to image space. Moreover, while DIP fixed the network input $z$ and trained the network parameters, here we keep the pre-trained diffusion model frozen and optimize with respect to input $z$ alone. The optimization objective is then equivalent to
\begin{align}\label{eq:p3}
    z^* = \arg \min_{z } \|g(z) - y\|^2, \quad\text{with}\quad x^* = g(z^*),
\end{align}
where we also let $E(g(z); y)=\|g(z) - y\|^2$. In general, the degradation can either \textit{remove} high-frequency details (resulting in smoothing) or \textit{add} high-frequency artifacts (\eg, noise). Here, we empirically investigate the behavior of the optimization described in \cref{eq:p3} under synthetic degradations that either add high-frequency artifacts (Gaussian noise) or remove high-frequency details (Gaussian blur). For this study, we randomly select images from the FFHQ dataset and apply: (1) Gaussian noise, and (2) Gaussian blur. We frame this as \textbf{a pure reconstruction task} and examine how the optimization in \cref{eq:p3} behaves when reconstructing the degraded images versus the original reference (clean) images. Our ultimate goal is to generalize these findings to more complex degradations, such as JPEG artifacts and non-uniform blur.
We run the optimization until it converges (this occurs with $N=1500$ iterations). In the case of a diffusion model, the mapping $g(z)$ could be computationally demanding, and this could make the iterative reconstruction in \cref{eq:p3} unfeasible. To enable this optimization, we adopt the recent approach presented in \cite{chihaoui2024blind}, where the inversion process is optimized for efficiency. However, in contrast to \cite{chihaoui2024blind} our final objective is not to use the inversion to reconstruct the clean image, but rather the degraded one. Also, unlike \cite{chihaoui2024blind} we do not use any degradation model. The detailed algorithm for our optimization is provided in Section \ref{algo22} of the appendix.
Our findings are that the answer to Q1 is positive and the answer to Q2 can be split into the following two observations
\begin{enumerate}
 \item \textit{There are two distinct regimes observed during the reconstruction, regardless of degradation type.} Interestingly, a key difference when using a pre-trained diffusion model compared to vanilla DIP is that, irrespective of the type of degradation (noise or blur), the optimization consistently produces clean images at an intermediate stage, even though it is purely a reconstruction process with \textbf{no integrated degradation model}. Specifically, there are two distinct regimes observed: (I) an initial regime where the generated images are clean, realistic, and progressively approach the clean reference images, followed by (II) a regime where the generated images start to reflect the degradation, becoming either blurred or noisy. This behavior is illustrated in Figure~\ref{fig:intermediate-steps}, by showing an example of intermediate reconstructions obtained during our optimization and that of DIP. In the case of DIP, in contrast, the reconstruction path is not consistent across the degradations. DIP fails to produce sharp and clean images during the iterative reconstruction in the case of blurry images.
\item \textit{There is a high inertia to reconstructing high-frequency artifacts (\eg, from noise).} One commonality observed between DIP and our proposed optimization scheme is the tendency to resist overfitting to high-frequency artifacts (such as noise). 
This behavior is illustrated in Figure~\ref{fig:intermediate-steps}, where the intermediate image remains clean until a late stage of the optimization (iteration = 500) when reconstructing an image corrupted with Gaussian noise.
\end{enumerate}

A crucial requirement for leveraging the findings from the previous section is an automated method to determine when to stop the optimization.
In the next paragraphs, we outline two simple and effective procedures to do so.


\begin{figure*}[t]
    \setkeys{Gin}{width=0.7\linewidth}
    \captionsetup[subfigure]{skip=0.5ex,
                             belowskip=1ex,
                             labelformat=simple}
    \renewcommand\thesubfigure{}

\subfloat[\centering $(a)$]{\includegraphics[width=0.5\textwidth]{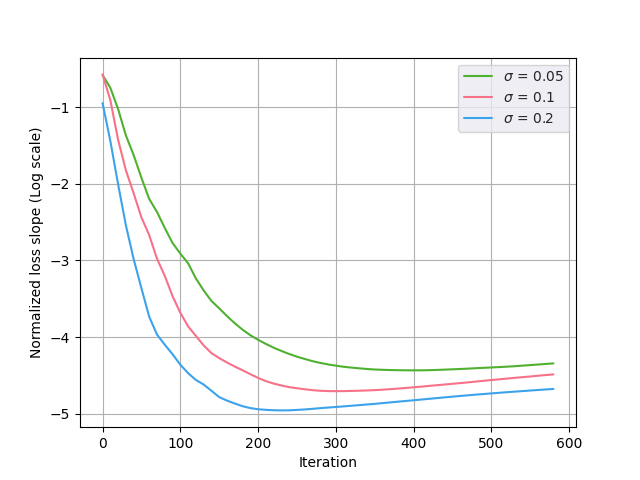}}
\hfill
\subfloat[\centering $(b)$
]{\includegraphics[width=0.5\textwidth]{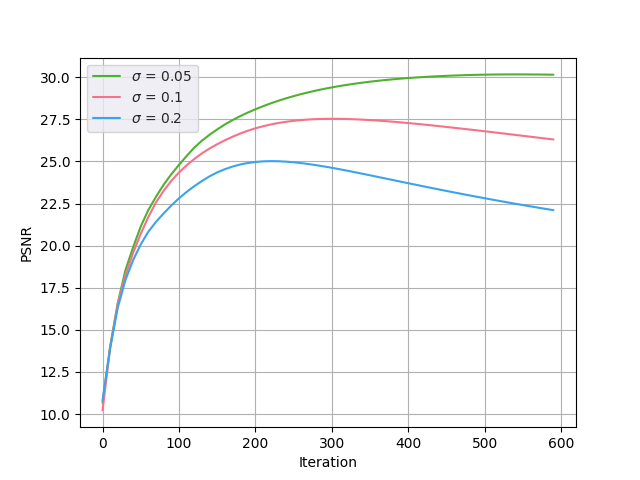}}
\hfill
    \caption{\label{fig:stopping-noise} \textbf{Stopping criterion in the case of degradations that introduce high-frequency artifacts:} We study the optimization trend across varying levels of degradation that introduce high-frequency artifacts, focusing specifically on input images with different noise levels. (a) Shows the normalized slopes of the loss functions on input images with increasing noise levels. (b) Shows the Peak Signal-to-Noise Ratio (PSNR) relative to the clean reference image as the optimization progresses for various noise intensities. We observe that the slopes in (a) reach their minima in correspondence to iterations where the corresponding PSNR in (b) is maximal. 
    For noise-based degradations (which introduce high-frequency artifacts), the slope in (a) 
    can then serve as an indicator of peak performance, providing guidance on the optimal stopping point for the optimization process.}
\end{figure*}

\subsection{Self-Supervised Stopping Criteria}

\subsubsection{Degradations removing high-frequency content}

\begin{figure}
    \centering
    \includegraphics[width=1.\linewidth]{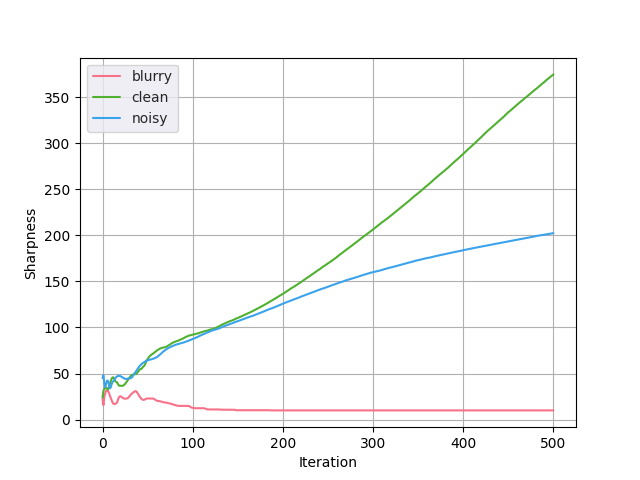}
    \caption{\textbf{Stopping criterion for degradations that remove high-frequency details from the image}. Here we show the trend of the variance of the Laplacian operator throughout our optimization scheme. After the first regime, in case of blurry image the variance of the Laplacian keeps decreasing while in the case of a sharp image the optimization adds more high frequency details and the variance of the Laplacian does not decrease. A similar trend to the clean image is observed for a noisy image, which shows that a separate stopping criterion is needed.}
    \label{fig:stopping-blur}
\end{figure}

Identifying an absolute measurement to characterize image sharpness is challenging. However, it is more manageable to assess \textbf{the relative change in the sharpness} of intermediate reconstructions. In our first finding, we observed that during regime (I) the generated images are realistic and sharp. Therefore, as the optimization progresses, we can detect when intermediate reconstructions begin to lose sharpness by using the variance of the Laplacian of the reconstructed image as a sharpness score. In Figure~\ref{fig:stopping-blur}, we illustrate the trend of the variance of the Laplacian (LV) operator throughout our optimization process. In the case of a blurry image, aside from a transient phase in the initial iterations, the variance steadily decreases as the image becomes less sharp, indicating a transition to overfitting. We define a minimum number of iterations $k_{min}$ to avoid the transitory regime and we keep track of the iteration number that corresponds to the last peak of sharpness. If the iteration  $k>k_{min}$ and  $\sigma^2[k+1] < \sigma^2[k]$, where $\sigma^2[k]\doteq LV(g(z^k))$, we stop the optimization. In this case, the optimal reconstruction is not necessarily at the current iteration ($k$), rather the one with the last highest sharpness (\ie, the highest LV). 

\subsubsection{Degradations adding high-frequency content}
In Figure~\ref{fig:stopping-noise}, we display the \emph{normalized slope} \begin{align}
\Delta_k = \frac{E(z^{k};y)- E(z^{k-1};y)}{E(z^{k-1};y)},     
\end{align}
where $k$ is the iteration number, in log scale (a) and the PSNR trend (b) relative to the clean reference image for input images with varying noise levels. 
We can immediately notice that the locations at which the slope $\Delta$ in (a) reach a minimum correspond to the maxima of the reconstruction in (b). 
Thus, a self-supervised stopping criterion based on the normalized slope of the loss can effectively detect when to halt optimization, maximizing reconstruction quality before overfitting occurs. For simplicity and robustness, our stopping criterion $detected_H$ is triggered (\textbf{true}) when the normalized loss decrease is below a threshold $\epsilon$. Formally, when $\Delta_k < \epsilon$, we stop and return the reconstruction $g( z^{k})$.




\begin{algorithm}[t!]
 \algsetup{linenosize=\small}
  \caption{\methodnameacr{}\label{alg:diip}}
  \begin{algorithmic}[1]
  \REQUIRE Degraded image $y$, $g$,$\eta$, $k_{min}$, $\epsilon$
  \ENSURE Restored image $\hat{x}$\\
  \STATE $k=0$, $LastSharpnessPeakIter=0$,\\ $detected_H$ = \FALSE, $detected_L$ = \FALSE\\ 
  \WHILE{NOT $detected_H$ \AND NOT $detected_L$}
      \STATE $z^{k+1} = z^{k} - \eta \nabla_{z} \|g( z^k) - y\|^2$
    
      \STATE Compute the Laplacian variance $\sigma^2[k]$ of $g( z^k)$ 
       \STATE \textit{\color{blue}//keep track of the iteration with last sharpness peak}
       \IF {  
       $\sigma^2[k]<\sigma^2[k-1]$ and
       $\sigma^2[k-2]<\sigma^2[k-1]$
       }
    \STATE $LastSharpnessPeakIter=k-1$ 
\ENDIF
 \STATE \textit{\color{blue}//Low frequency degradation stopping criterion}
  \IF { $k > k_{min}$ and $\sigma^2[k]<\sigma^2[k-1]$ 
  }
    \STATE $detected_L$ = \TRUE
     \STATE $n^*=LastSharpnessPeakIter$ 
\ENDIF
 \STATE \textit{\color{blue}//High frequency degradation stopping criterion}

  \IF { $k > \tau$ and $\Delta_k < \epsilon$ 
  }
    \STATE $detected_H$ = \TRUE
     \STATE $n^*=k-\tau$
\ENDIF
      
\ENDWHILE
 
  \STATE Return the restored image $\hat{x}=g(z^{n^*})$\\
\end{algorithmic}

\end{algorithm}

\subsection{Restoration via the \methodName}

Based on the previous study, we propose a simple algorithm for image restoration. Given a degraded image \( y \), we run the optimization described in \cref{eq:p3} using gradient descent with a random initial $z$.
At each iteration, we take a gradient descent step and check if either of the stopping criteria about high-frequency artifacts or low-frequency artifacts is met. To address low-frequency artifacts, we monitor the trend of the Laplacian variance. If we reach a minimum iteration and if the Laplacian variance is decreasing, we stop the optimization and we return the intermediate prediction that corresponds to the last peak of the Laplacian variance. To address high-frequency artifacts, we monitor the loss reduction. If the normalized slope of the loss decreases below a certain threshold $\epsilon$, we terminate the optimization process.  Our algorithm is outlined in Algorithm~\ref{alg:diip}.





\begin{figure*}[t]
\centering\
   \setkeys{Gin}{width=0.16\linewidth}
    \captionsetup[subfigure]{skip=0.0ex,
                             belowskip=0.0ex,
                             labelformat=simple}
                             \setlength{\tabcolsep}{0.0pt}
    \renewcommand\thesubfigure{}
    \small

\small
  \begin{tabular}{cccccc}
  \textit{Input} & \textit{BlindDPS} & \textit{BIRD} & \textit{DreamClean} & \textit{\methodnameacr} & \textit{Target} \\
 {\includegraphics{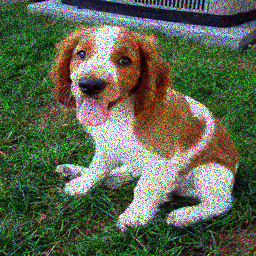}}& {\includegraphics{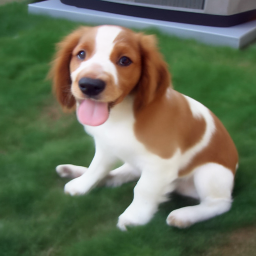}}& {\includegraphics{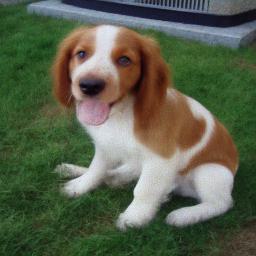}}& {\includegraphics{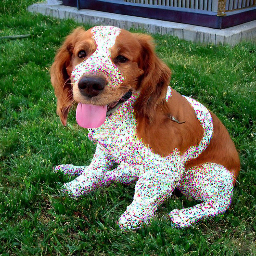}} & 
 {\includegraphics{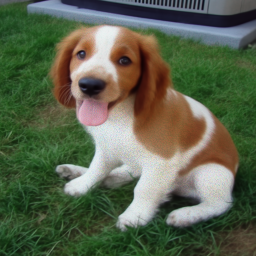}} & 
 {\includegraphics{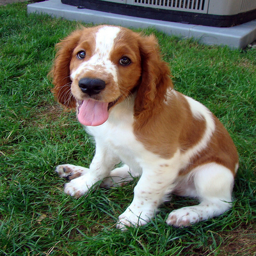}}\\
  {\includegraphics{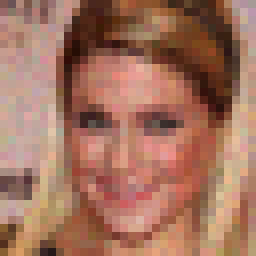}}& {\includegraphics{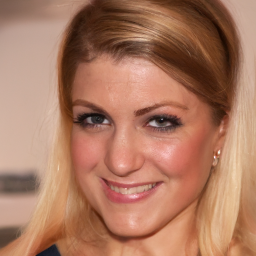}}& {\includegraphics{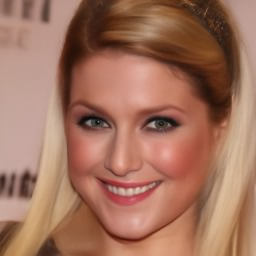}}& {\includegraphics{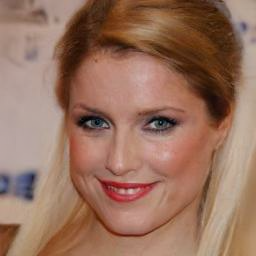}} & 
 {\includegraphics{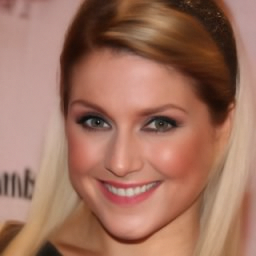}} & 
 {\includegraphics{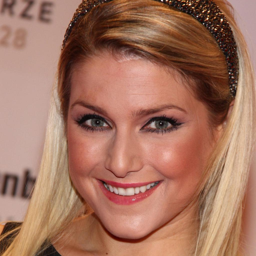}}\\
 
   \end{tabular}

\captionof{figure}{\label{fig:celeba} Qualitative comparisons of blind methods on restoration tasks with structured degradations: Top: Denoising. Bottom: Super-resolution (×4). \vspace{.5em}}
\end{figure*}

\vspace{-0.2cm}
\section{Experiments}

\begin{table*}[t]
\centering
\caption{ Quantitative comparison with training free and zero-shot blind zero-shot methods on structured IR tasks on the CelebA validation dataset. The best method is indicated in bold.\label{tab:partially-blind}}
     \centering
    \footnotesize
    \captionsetup{font=tiny}
\begin{tabularx}{\textwidth}
{@{\extracolsep{\fill}}lccccccccc
}
\toprule
Method & \multicolumn{3}{c}{Denoising}  & \multicolumn{3}{c}{ Superresolution ($\times 4$)}  & \multicolumn{3}{c}{ Superresolution ($\times 8$)}\\ 
     & PSNR $\uparrow$ & SSIM $\uparrow$  & LPIPS $\downarrow$  & PSNR $\uparrow$ & SSIM $\uparrow$  & LPIPS $\downarrow$ & PSNR $\uparrow$  & SSIM $\uparrow$ & LPIPS $\downarrow$\\ \midrule

     GDP~\cite{fei2023generative}   & {27.73}   & {0.817}       &  {0.232} & {24.21}       &  {0.708} & {0.337}      &  21.66 & 0.618 & 0.374     \\  
         Gibbsddrm~\cite{murata2023gibbsddrm}      & {27.38}   & {0.809}       &  {0.255} & {24.38}       &  {0.689} & {0.330}      &  {21.45} & {0.605} & {0.364}      \\ 
    BIRD~\cite{chihaoui2024blind}    & {27.92}   & {0.821}       &  {0.238} & \textbf{25.26}       &  {0.751} & \textbf{0.294}      &  {22.63}& {0.626} & {0.352}      \\     
     
          BlindDPS~\cite{chung2023parallel}   & {27.56}   & {0.813}       &  {0.246}  & {24.51}       &  {0.722} & {0.324}      &  21.73 & 0.620 & 0.360      \\ 
            DIP~\cite{ulyanov2018deep}      &    25.81    & 0.606         &  0.345 & 21.33    &  0.566 & 0.426      &  20.34 & 0.488  & 0.471           \\
               DreamClean~\cite{ulyanov2018deep}      &    27.05    & 0.771         &  0.236 & 23.44    &  0.663 & 0.322      &  21.33 & 0.586  & 0.344           \\
      Ours     & \textbf{28.37}   & \textbf{0.842}       &  \textbf{0.224} & {25.14}       &  \textbf{0.764} & {0.301}      &  \textbf{22.86} & \textbf{0.651} & \textbf{0.336}     \\  \bottomrule 
\end{tabularx}
\end{table*}

\begin{table*}[ht]
\centering
\caption{Quantitative comparison with training free and zero-shot blind  methods on unstructured IR tasks. We note that\cite{chung2023parallel, fei2023generative, chihaoui2024blind, murata2023gibbsddrm} could not be applied to unstructured degradations. The best method is indicated in bold.\label{tab:fully-blind}}
     \centering
    \footnotesize
    \captionsetup{font=tiny}
\begin{tabular*}{\textwidth}{@{\extracolsep{\fill}}lccccccccr
}
\toprule
Method & \multicolumn{3}{c}{JPEG De-artifacting}  & \multicolumn{3}{c}{Non-uniform Deformation}& \multicolumn{3}{c}{Water-drop Removal}\\ 
     & PSNR $\uparrow$ & SSIM $\uparrow$  & LPIPS $\downarrow$  & PSNR $\uparrow$ & SSIM $\uparrow$  & LPIPS $\downarrow$ & PSNR $\uparrow$ & SSIM $\uparrow$  & LPIPS $\downarrow$\\ \midrule
       DIP~\cite{ulyanov2018deep}      &   20.43    & 0.593         &  0.622 &  18.83    & 0.437         &  0.643 & 20.37    &  0.517 & 0.642           \\
     DreamClean~\cite{xiaodreamclean}      & {23.92}   & {0.691}       &  {0.342} & {22.16}       &  {0.612} & {0.398}  & 22.94   &  0.643 & \textbf{0.361}         \\ 
      Ours    & \textbf{25.29}   & \textbf{0.783}       &  \textbf{0.325} &  \textbf{23.45} & \textbf{0.689}    & \textbf{0.392}           & \textbf{23.78}    &  \textbf{0.702} & {0.377}   \\  \bottomrule 
\end{tabular*}
\end{table*}

\noindent\textbf{Experimental Settings. }
We evaluate \methodnameacr across different IR tasks  on CelebA~\cite{liu2018large}, ImageNet~\cite{deng2009imagenet}. We use unconditional pre-trained diffusion models trained where the backbone is a Unet~\cite{ronneberger2015u} architecture. For a fair comparison, we use the same pre-trained models across all compared methods.  
For quantitative evaluation, we conduct experiments on classic image restoration tasks, including structured degradations (\eg, Denoising, super-resolution) and complex unstructured degradations (\eg, non-uniform deformation, water drop removal). 
We evaluate our method on ImageNet 1K, CelebA 1K using images of size \(256 \times 256\) pixels. Performance is measured using Peak Signal-to-Noise Ratio (PSNR) and Structural Similarity Index Measure (SSIM) for fidelity assessment, while Learned Perceptual Image Patch Similarity (LPIPS) is used as a perceptual metric. \methodnameacr is compared against state-of-the-art zero-shot and partially-blind methods, including BlindDPS~\cite{chung2023parallel}, GDP~\cite{fei2023generative}, BIRD~\cite{chihaoui2024blind} and GibbsDDRM~\cite{murata2023gibbsddrm} . Additionally, we compare against fully blind zero-shot methods such as DIP~\cite{ulyanov2018deep} and the recent state-of-the-art DreamClean~\cite{xiaodreamclean}. We would like to emphasize that only  a few methods have been proposed for the case of unstructured degradations (\eg, water drop removal), which explains the smaller number of methods compared in Table~\ref{tab:fully-blind} relative to Table~\ref{tab:partially-blind}.  
 For super-resolution, we apply an \(8 \times 8\) Gaussian kernel followed by an \(8\times\) downsampling operation. For image denoising, we use a mixture of Gaussian and speckle noise with $\sigma \simeq0.3$ For JPEG artifact removal, we generate degraded inputs using the Imageio~\cite{chandra2001imageio} Python library with a quality factor of \(q=5\). In all tasks, we consider the presence of additive noise with a standard deviation of \(\sigma=0.02\). For the generation of unstructrued degradations, we use an online tool\footnote{https://online.visual-paradigm.com/} to generate degraded inputs.  
 We set $k_{min}=100$ and $\epsilon=0.001$. Optimization is performed using the Adam optimizer with a learning rate of $0.0015$. 

\begin{table}[t]
  
\centering
\caption{Effect of $k_{min}$ on  \methodnameacr performance. PSNR (dB) is reported when using  different $k_{min}$ values. \label{tab:k-min}}
     \centering
    \footnotesize
    \captionsetup{font=scriptsize}
\begin{tabular*}{\linewidth}{
@{\extracolsep{\fill}}lcc
}
\toprule
$k_{min}$     & Non-uniform deformation & Waterdrop removal  \\ \midrule
$50$    & 22.18 &  22.48\\
$100$    &      23.45 & 23.78            \\
$150$    &      23.52 & 23.82          \\

\bottomrule
\end{tabular*}
\end{table}

\begin{table}[ht]
  
\centering
\caption{Effect of $\epsilon$ on  \methodnameacr performance. PSNR (dB) is reported when using  different $\epsilon$ values. \label{tab:epsilon}}
     \centering
    \footnotesize
    \captionsetup{font=scriptsize}
\begin{tabular*}{\linewidth}{
@{\extracolsep{\fill}}lcc
}
\toprule
$\epsilon$     & Denoising & JPEG-deartifacting  \\ \midrule
$0.005$    & 27.25 &  22.38\\
$0.001$    &      28.37 & 25.29          \\
$0.0005$    &      28.14 & 25.02          \\

\bottomrule
\end{tabular*}
\end{table}

\begin{table}[ht]
  
\centering
\caption{Performance gap when using optimal stopping. PSNR (dB) is reported for our stopping criterion and an optimal stopping criterion that has access to ground-truth data. \label{tab:stopping}}
     \centering
    \footnotesize
    \captionsetup{font=scriptsize}
\begin{tabular*}{\linewidth}{
@{\extracolsep{\fill}}lcc
}
\toprule
    & Denoising & Water-drop removal  \\ \midrule
 Optimal stopping   & 28.63 &  23.73\\
Ours   &      28.37 & 23.45            \\

\bottomrule
\end{tabular*}
\end{table}

\noindent\textbf{Quantitative and Qualitative Comparison.}
Tables~\ref{tab:partially-blind} and \ref{tab:fully-blind} show the performance of \methodnameacr compared to other methods in both the partially blind and fully blind settings. \methodnameacr consistently outperforms or matches the state-of-the-art approaches across different image restoration tasks. It is important to note that methods such as \cite{fei2023generative, chihaoui2024blind, murata2023gibbsddrm, chung2023parallel} are not applicable in the fully blind scenario, as they require knowledge of the parametric form of the degradation model, and therefore are not included in Table~\ref{tab:fully-blind}. 
Figure~\ref{fig:celeba} presents a visual comparison of CelebA images restored for image denoising and super-resolution tasks. Figure~\ref{fig:unstructured} showcases a visual comparison for unstructured restoration tasks. Despite lacking prior knowledge of the degradation, \methodnameacr produces realistic reconstructions and generally preserves higher fidelity compared to other competing methods.


 


\begin{table}[]
\centering
\caption{Runtime (in seconds) and Memory consumption (in Gigabytes) comparison of training-free methods on CelebA. The input image is of size $256 \times 256$. \label{tab:additional_comparisons_runtime}}
     \centering
    \scriptsize
    \captionsetup{font=scriptsize}
\begin{tabular*}{\linewidth}{
@{\extracolsep{\fill}}lcc
}
\toprule
Method     & Runtime [s] & Memory [GB]   \\ \midrule
 GDP~\cite{fei2023generative}     &       168  &       1.1              \\
BIRD~\cite{chihaoui2024blind}     &    234  &    1.2   \\
BlindDPS~\cite{chung2023parallel} & 270 & 6.1\\
DreamClean~\cite{xiaodreamclean} & 125 & 1.3\\

Ours    &      138 &      1.2         \\

\bottomrule
\end{tabular*}
\end{table}

\begin{figure}[t]
\centering\
   \setkeys{Gin}{width=0.2\linewidth}
    \captionsetup[subfigure]{skip=0.0ex,
                             belowskip=0.0ex,
                             labelformat=simple}
                             \setlength{\tabcolsep}{0.0pt}
    \renewcommand\thesubfigure{}
    \small

\small
  \begin{tabular}{ccccc}
  \textit{Input} & \textit{DIP} & \textit{DreamClean} & \textit{Ours} & \textit{Target} \\
 {\includegraphics{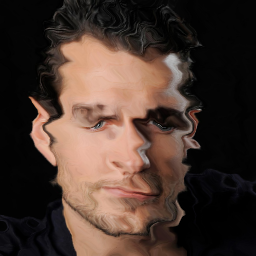}}& {\includegraphics{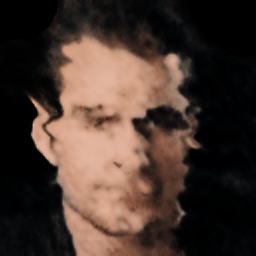}}& {\includegraphics{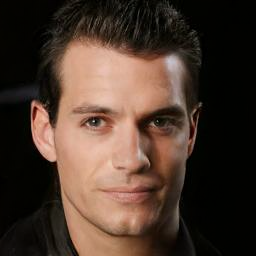}} & 
 {\includegraphics{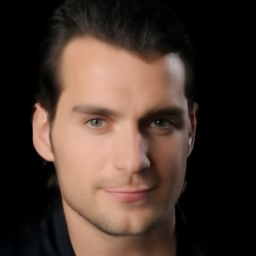}} & 
 {\includegraphics{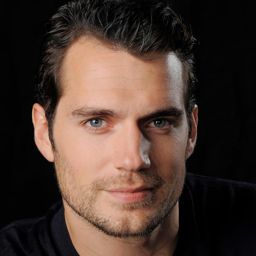}}\\
 {\includegraphics{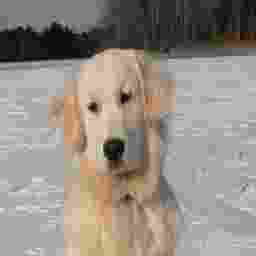}}& {\includegraphics{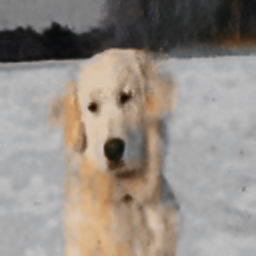}}& {\includegraphics{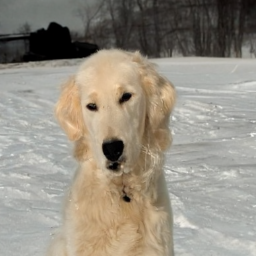}} & 
 {\includegraphics{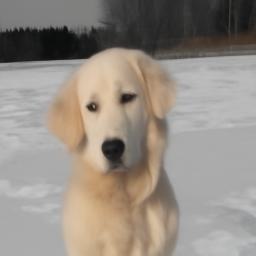}} & 
 {\includegraphics{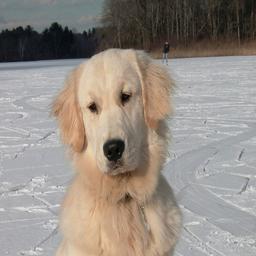}}\\
  {\includegraphics{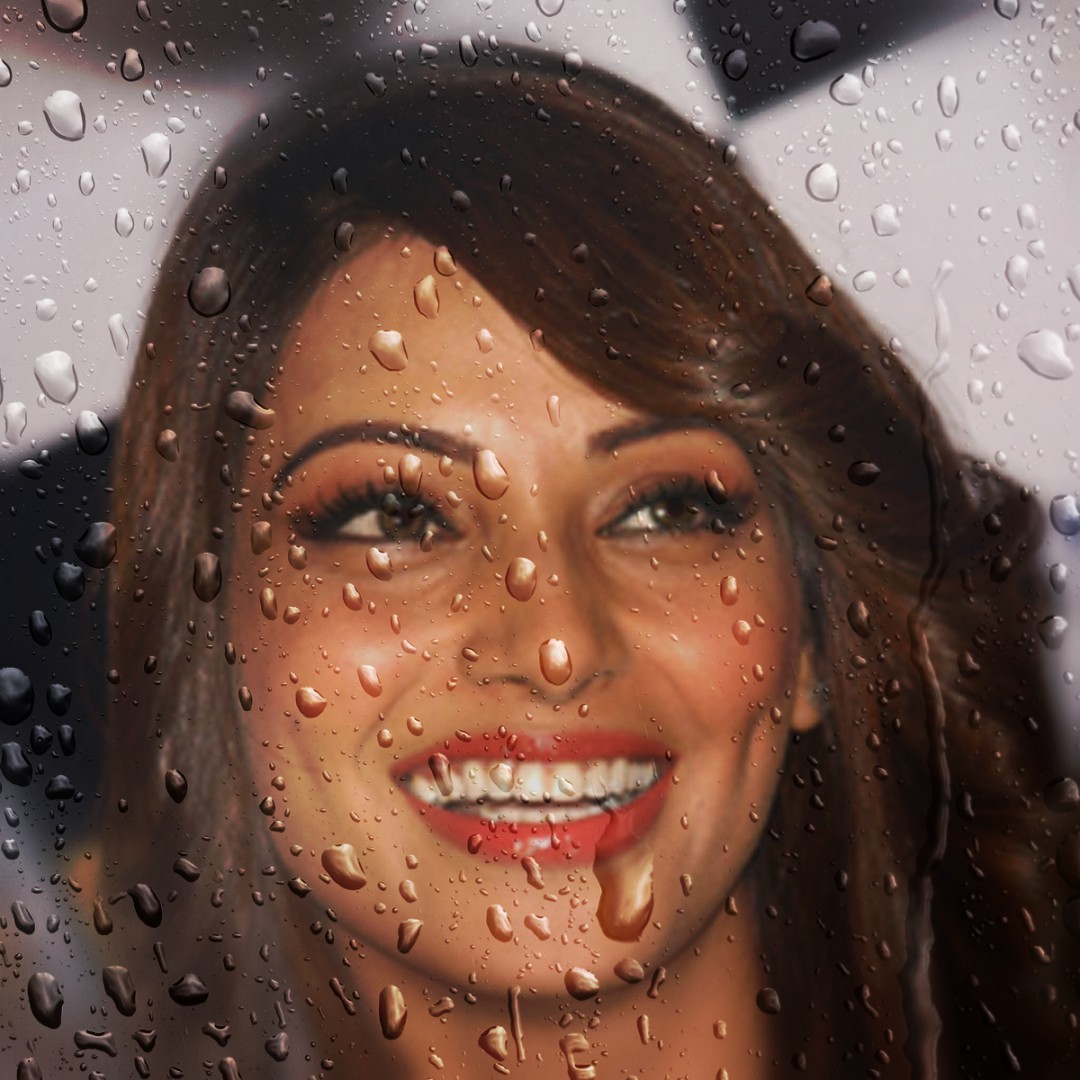}}& {\includegraphics{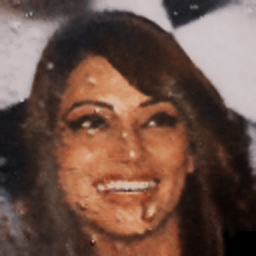}}& {\includegraphics{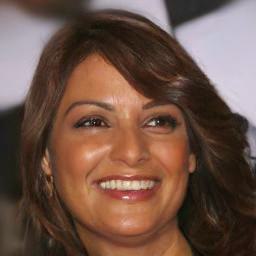}} & 
 {\includegraphics{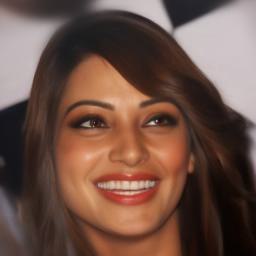}} & 
 {\includegraphics{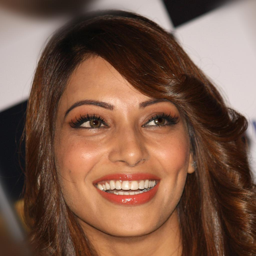}}\\
 
   \end{tabular}

\captionof{figure}{\label{fig:unstructured} Visual comparisons of blind methods for restoration tasks involving unstructured degradations.\vspace{.5em}}
\end{figure}

\section{Ablations}

We conduct different ablations to analyze the impact of each component of \methodnameacr. Specifically, we examine the influence of the two hyper-parameters, $k_{min}$ and $\epsilon$, in Algorithm~\ref{alg:diip}.

\noindent\textbf{Effect of the minimum number of iterations $k_{min}$.}
 In Table \ref{tab:k-min}, we present the effect of the minimum iteration \( k_{min} \) on \methodnameacr. After approximately \( k_{min} = 100 \), the performance stabilizes, making it a good trade-off between image quality and efficiency.

\noindent\textbf{Effect of the threshold $\epsilon$.}
 In Table \ref{tab:epsilon}, we present the effect of the threshold \( \epsilon \). A higher \( \epsilon \) causes earlier stopping compared to the optimal value, resulting in a lower PSNR. Conversely, a very small \( \epsilon \) allows the process to continue longer, leading to high-frequency artifacts. We found that \( \epsilon = 0.001 \) provides the best performance.  
 
\noindent\textbf{Gap to the optimal stopping.}
 In Table \ref{tab:stopping}, we quantify the gap between our stopping criterion and the optimal one, assuming access to the target clean image. To establish the optimal stopping point, we run a baseline using our optimization and stop when the PSNR to the ground truth ceases to improve. \methodnameacr lags behind the optimal stopping point by approximately $0.3$ dB in the cases of denoising and water drop removal.

\noindent\textbf{Efficiency Comparison.}
In Table \ref{tab:additional_comparisons_runtime}, we compare the runtime and memory consumption of training-free methods. We run \methodnameacr on $100$ degraded images with different degradation type and we report our as the average time. Our method achieves a good balance between image quality and computational efficiency while offering a broader range of applicability compared to most existing methods.

\section{Conclusion}

In this work, we presented the \methodName (\methodnameacr), a novel blind image restoration method that leverages pretrained diffusion models to handle a wide variety of degradation types without requiring explicit knowledge of the degradation process. Drawing inspiration from the Deep Image Prior (DIP), we demonstrated that pretrained diffusion models provide a much stronger prior for restoration tasks, enabling the model to effectively reconstruct clean images even when faced with complex and unknown degradations. Our experiments showed that the optimization process in \methodnameacr is capable of producing high-fidelity restored images across multiple degradation types, including JPEG artifact removal, waterdrop removal, non-uniform deformation, and super-resolution. By introducing a strategy of early stopping, we were able to avoid overfitting to the degraded input, further improving restoration quality.  \methodnameacr achieves competitive results in terms of both restoration quality and robustness to diverse degradations, offering a promising approach for real-world image restoration tasks where degradation models are unknown or too complex to define.

{
    \small
    \bibliographystyle{ieeenat_fullname}
    \bibliography{main}
}

\newpage
\appendix
\section{Algorithm to Solve \cref{eq:p3} \label{algo22}}

 Given a degraded image $y$ and a mapping $g$ induced by a pre-trained diffusion. We aim to solve the optimization objective:
\begin{align}\label{eq:p3}
    z^* = \arg \min_{z } \|g(z) - y\|^2, \quad\text{with}\quad x^* = g(z^*),
\end{align}

We adopt a similar strategy to that proposed in \cite{chihaoui2024blind} however in contrast to \cite{chihaoui2024blind}, no specific degradation model is assumed. Rather, the focus is solely on reconstructing the original degraded image. The details of the algorithm are provided in \cref{alg:d2d}. Similarly to \cite{chihaoui2024blind}, we set $\delta t = 100$.

\begin{algorithm}[H]
 \algsetup{linenosize=\small}
  \scriptsize
  \caption{Algorithm to Solve \cref{eq:p3} \label{alg:d2d}}
  \begin{algorithmic}[1]
  \REQUIRE Degraded image $y$, learning rate $\eta$, step size $\delta t$, threshold $\epsilon$, diffusion steps $T$.
  \ENSURE Return $\tilde {z}$\\

 \STATE Initialize $z  \sim \mathcal{N}(\mathbf{0}, \mathbf{I})$, $\hat{x}_{0}=\mathbf{0}$
  \WHILE{$\|\hat{x}_{0} - y\| > \epsilon$}  

   \STATE $t = T, z_T = z$ 
  \WHILE{$t > 0$}  
    \STATE $\hat{x}_{0} = (z_t - \sqrt{1 - \bar{\alpha}_{t}}  \epsilon_{\theta} (z_t, t))/\sqrt{\bar{\alpha}_{t}}$ 
 
  \STATE $z_{t- {\delta t}} = \sqrt{\bar{\alpha}_{t-{\delta t}}}  \hat{x}_{0} + \sqrt{1 - \bar{\alpha}_{t-{\delta t}}} . \frac{z_t - \sqrt{\bar{\alpha}_{t}} \hat{x}_{0}}{\sqrt{1 - \bar{\alpha}_{t}}}$
 
   \STATE $t \leftarrow t - \delta t$
  \ENDWHILE\\
  \STATE $z \leftarrow z - \eta \nabla_{z} \|\hat{x}_{0} - y\|$
  
  \ENDWHILE\\
  \STATE \textbf{return} $ {z}$
\end{algorithmic}

\end{algorithm}






\section{Additional Visual Comparisons}

We include additional visual comparisons in Figure \ref{fig:imagenet}. 




\begin{figure*}[t]
\centering\
   \setkeys{Gin}{width=0.25\linewidth}
    \captionsetup[subfigure]{skip=0.0ex,
                             belowskip=0.0ex,
                             labelformat=simple}
                             \setlength{\tabcolsep}{1.5pt}
    \renewcommand\thesubfigure{}
    \small

\small
  \begin{tabular}{cccc}
  \textit{Input} & \textit{DreamClean}~\cite{xiaodreamclean} & \textit{\methodnameacr} & \textit{Ground-truth} \\
 {\includegraphics{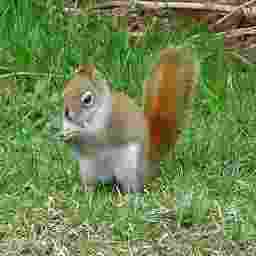}}& {\includegraphics{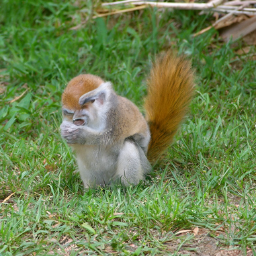}}& {\includegraphics{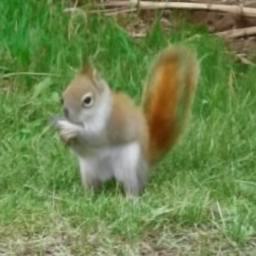}}& 
{\includegraphics{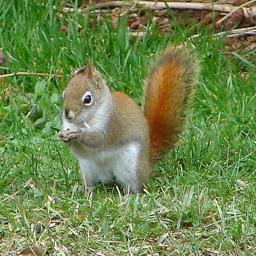}}\\{\includegraphics{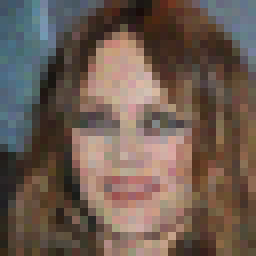}}& {\includegraphics{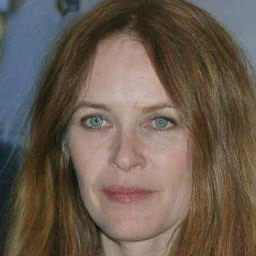}}& {\includegraphics{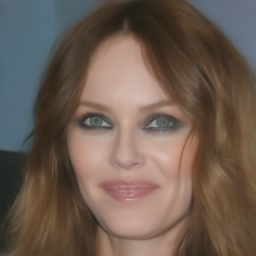}}& 
{\includegraphics{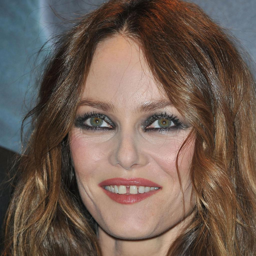}}\\
 {\includegraphics{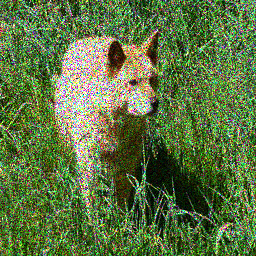}}& {\includegraphics{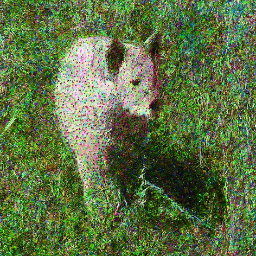}}& {\includegraphics{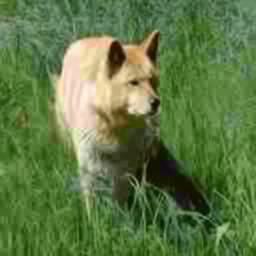}}& 
{\includegraphics{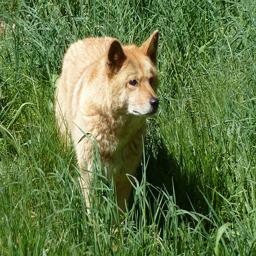}}
   \end{tabular}

\captionof{figure}{\label{fig:imagenet} Visual comparisons . Top: JPEG de-artifacting. Middle: superresolution ($\times 4$). Bottom: blind denoising with a mixture of Gaussian and speckle noise.\vspace{.5em}}
\end{figure*}

\end{document}

%% file: preamble.tex
\newcommand{\methodnameacr}{DIIP\xspace}
\newcommand{\methodName}{DIffusion Image Prior\xspace}